\documentclass[10pt,twocolumn,letterpaper]{article}

\usepackage{iccv}
\usepackage{times}
\usepackage{epsfig}
\usepackage{graphicx}
\usepackage{amsmath}
\usepackage{amssymb}

\usepackage{mathtools}
\usepackage{multirow}
\usepackage{enumitem}
\usepackage{color}
\def\red#1{\textcolor[rgb]{1,0,0}{#1}}

\newcommand{\keypoint}[1]{\vspace{0.1cm}\noindent\textbf{#1}\quad}
\newcommand{\cut}[1]{}
\usepackage[nosort, nocompress]{cite}
\usepackage{algorithm}
\usepackage{algpseudocode}
\usepackage{makecell}
\usepackage{lipsum}
\usepackage[pagebackref=true,breaklinks=true,letterpaper=true,colorlinks,bookmarks=false]{hyperref}

\iccvfinalcopy 

\def\iccvPaperID{3747} 

\ificcvfinal\fi

\begin{document}

\title{Towards the Unseen: Iterative Text Recognition by Distilling from Errors}

\author{Ayan Kumar Bhunia\textsuperscript{1} \hspace{.2cm}  Pinaki Nath Chowdhury\textsuperscript{1,2}\hspace{.2cm}  Aneeshan Sain\textsuperscript{1,2}\hspace{.2cm} Yi-Zhe Song\textsuperscript{1,2} \\
\textsuperscript{1}SketchX, CVSSP, University of Surrey, United Kingdom. 
\\
\textsuperscript{2}iFlyTek-Surrey Joint Research Centre
on Artificial Intelligence. 
\\{\tt\small \{a.bhunia, p.chowdhury, a.sain, y.song\}@surrey.ac.uk}.  
}


\maketitle
\ificcvfinal\thispagestyle{empty}\fi

\begin{abstract}
Visual text recognition is undoubtedly one of the most extensively researched topics in computer vision. Great progress have been made to date, with the latest models starting to focus on the more practical ``in-the-wild'' setting. However, a salient problem still hinders practical deployment -- prior state-of-arts mostly struggle with recognising unseen (or rarely seen) character sequences. In this paper, we put forward a novel framework to specifically tackle this ``unseen'' problem. Our framework is iterative in nature, in that it utilises predicted knowledge of character sequences from a previous iteration, to augment the main network in improving the next prediction. Key to our success is a unique cross-modal variational autoencoder to act as a feedback module, which is trained with the presence of textual error distribution data. This module importantly translate a discrete predicted character space, to a continuous affine transformation parameter space used to condition the visual feature map at next iteration.
Experiments on common datasets have shown competitive performance over state-of-the-arts under the conventional setting. Most importantly, under the new disjoint setup where train-test labels are mutually exclusive, ours offers the best performance thus showcasing the capability of generalising onto unseen words (Figure~\ref{fig:Fig1} offers a summary).
\end{abstract}

\vspace{-0.5cm}
\section{Introduction}

Text recognition being a longstanding problem in computer vision plays a pivotal role in a diverse set of applications, ranging from OCR systems \cite{bhunia2019handwriting, poznanski2016cnn, Shi2017, wang2020scene}, navigation and guiding board recognition \cite{AONRecogCVPR2018}, to more recent ones such as visual question answering \cite{biten2019scene}. With the advance of deep learning \cite{ShiBaiPAMI2019, ESIR2019, AONRecogCVPR2018, qiao2020seed}, recognition accuracy have notably increased over traditional methods \cite{Neumann2012}. Research focus has thus shifted to the more practical ``in-the-wild'' setting in an attempt towards ubiquity. Of these, irregular scene text recognition \cite{ShiBaiPAMI2019, AONRecogCVPR2018, yang2019symmetry, yue2020robustscanner} has gained considerable attention, yet the focus is placed on irregular-image rectification process \cite{ESIR2019, yang2019symmetry} other than the core recognition problem itself.

\begin{figure}[t]
\begin{center}
  \includegraphics[width=0.95\linewidth]{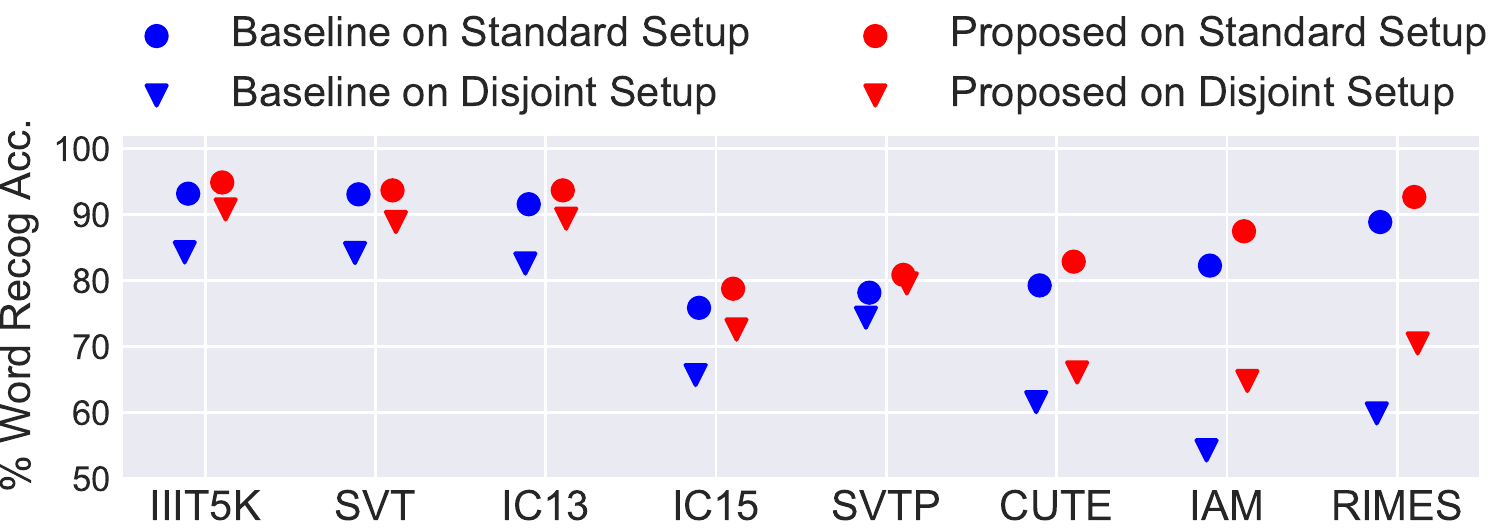}
\end{center}
\vspace{-0.35cm}
  \caption{
 Figure shows how the performance of baseline model \cite{ShiBaiPAMI2019}
 is limited under disjoint setup where testing words are not encountered during training. Our method performs way better than the baseline in the disjoint setup, reducing the performance gap with standard setup and showcasing its potential onto rarely seen words. Nevertheless, improvement in standard setup can be noticed over all datasets as well.}
\label{fig:Fig1}
\vspace{-0.6cm} 
\end{figure}

In this paper, we continue this push towards practicality, albeit with a different perspective -- we importantly focus on the understudied problem of unseen (or rarely seen) word recognition, where no (or limited) word image of a particular character sequence is present during training. Our motivation is straightforward -- humans can recognise a word image, even when it falls beyond the scope of known vocabulary. In fact, robustness of a text recognition framework largely depends on its performance on rarely or unseen words \cite{wan2020vocabulary}. {Note that unlike the conventional zero-shot \cite{ye2017zero} setting  where the transfer happens on class-level, here the combination of characters is ``unseen" although the characters individually have come up in training. The fact that the sequences not being encountered during training is what makes this task challenging.} Our solution for this ``unseen'' problem is intuitive: (i) we leverage an iterative framework with a feedback mechanism to give the model a chance to re-visit its false predictions, and (ii) we explicitly ask such feedback to encapsulate useful information that would help the model to correct itself at the following iteration.

Our first contribution is therefore an iterative framework, where characters predicted in the previous iteration provide clues through a feedback loop \cite{HuhZhangCVPR2019} to enhance performance in the subsequent iterations. This is fundamentally different to current state-of-the-arts \cite{ShiBaiPAMI2019, AONRecogCVPR2018, litman2020scatter}, most of which adapt a feed-forward framework consisting of three-components (feature-extracting backbone, bidirectional-RNN encoder, and attention-based recurrent decoder). 
Despite the attention mechanism, its \textit{single-pass} nature still dictates wrong predictions, thereby leaving no chance for the model to recover.
To this end, our iterative design enables the revision of incorrect intermediate predictions in its subsequent steps, via a novel cross-modal (i.e., text prediction to image feature-maps) feedback mechanism. 
The key to our success lies with \textit{how} feedback is progressed at each iteration. 
A naive solution might be to apply an independent spelling correction network~\cite{GuoWeiss2019, XieNG2016} chained serially to a basic text recognition model. Apart from not being end-to-end trainable, this also ignores the intermediate \textit{visual} features from the recognition network, ultimately bisecting the feedback loop. 
We on the other hand advocate that earlier word predictions (text labels) should be fed back {cross-modal} to the main text recognition network and directly modulate the visual feature maps at the next iteration. That is, the feedback module triggers a mapping from the discrete predicated label space, to a continuous space of affine transformation parameters (akin to \cite{perez2018film}) which are consequently used to condition visual features (hence closing the feedback loop).

Simply knowing how feedback works is not enough -- we still need to devise \textit{what} information should be fed back to give a model its best shot at rectifying itself. For this, {we resort to distilling knowledge from textual error distributions collected from state-of-the-art text recognition models} -- this is akin to humans who use prior experiences to help them to make corrections. For example, `$hello$' might be wrongly predicted as `$nello$' or `$bello$' due to partial structural similarity of `$h$' with `$n$' or `$b$'. By distilling such {error distributions} into the feedback module (\textit{during training only}), the model will gain knowledge of correct character associations. Our second contribution is therefore designing the feedback module via a conditional variational auto-encoder (CVAE) \cite{sohn2015learning} that learns from such error distributions. More specifically, we augment the vanilla CVAE with an \emph{auxiliary decoder} that tries to directly reconstruct the correct word, given any incorrect prediction at each iteration. Note that deterministic alternatives such as typical feedback networks \cite{ShamaManorAFL2018, HuhZhangCVPR2019} or spelling correction (prediction refining) networks, \cite{GuoWeiss2019, XieNG2016} would not work well since they do not model the uncertainty among multiple erroneous alternatives, dictating a variational formulation like ours.

Our contributions are:  [a] We for the first time propose an \emph{iterative approach} to specifically tackle the ``unseen'' text recognition problem. [b] We design a conditional variational autoencoder to act as a feedback module, which works \textit{cross-modal} to propagate predicted \textit{text labels} from an earlier iteration to condition the \textit{visual features} from the main network.  [c] Our novel cross-modal feedback module is trained by distilling knowledge learned from \emph{textual error distributions} that model multiple erroneous character sequences to a given candidate word. 

Experiments confirm our framework to be capable of adopting unseen words better than state-of-the-art frameworks on various public scene-text recognition and handwriting recognition datasets. Further ablative studies demonstrate the superiority of our iterative framework over naive spell checking \cite{XieNG2016, GuoWeiss2019} and language model alternatives \cite{kang2019candidate}, and that the proposed feedback module can be plug-and-play with more than one base network.

\vspace{-0.15cm}
\section{Related Works}
\vspace{-0.15cm}
\noindent \textbf{Text Recognition:}
\cut{An overview of some traditional text reading methods are discussed in depth by Ye \etal \cite{ye2014text}.} With the rising applicability of deep learning methods, \cut{Bissacco \etal \cite{BissacoICCV2013} and} Jaderberg \etal \cite{JaderbergARXIV2014} employed convolution neural network for text recognition, but such methods were constrained to dictionary words. Although this limitation was eliminated by Jaderberg \etal \cite{Jaderberg2015}, it still needed huge resources for character level localisation. Sequence discriminative training using connectionist temporal classification (CTC) \cite{GravesICML2006} coupled with recurrent neural networks, dealt with the need for character level localisation. This lead to an end-to-end trainable convolutional recurrent neural network for reading texts \cite{Shi2017}. This was further enhanced by incorporating the idea of attention \cite{BahdanauBengio2014} for text recognition \cite{LeeCVPR2016,ShiBai2016}. Usually a two-fold process is implemented \cite{yang2019symmetry, ShiBai2016, ShiBaiPAMI2019, ESIR2019}, where an irregular image is first passed through a rectification network, and then followed by a text recognition network. Ideas like 2D attention mechanism, focusing attention network (FAN) \cite{ChengICCV2017} has been explored recently along-with the possibility of enhancing text reading accuracy using synthetically generated large datasets \cite{VeriSimilarECCV2018}. Despite such extensive research, reading irregular curved texts had not been explored much in details until recently in \cite{AONRecogCVPR2018}, which describes how arbitrarily oriented texts can be read by extracting four directional features from the 2D input image. Baek \etal \cite{baek2019wrong} has conducted a comparative study where different popular text recognition architectures were trained in a similar setting. Hence it can be observed that although recent works in text reading have emphasised on designing a better rectification network \cite{ESIR2019, yang2019symmetry}, all such researches \cite{luo2020learn, xu2020machines} essentially used an off-the-shelf \cite{Shi2017} recognition module. On the other side, handwriting recognition poses a tougher challenge owing to a free flow nature \cite{bhunia2019handwriting} of writing. Poznanski \etal \cite{poznanski2016cnn} employed a ConvNet to estimate an n-gram frequency profile along with a large dictionary having true frequency profiles for recognition. 

\vspace{0.1cm}
\noindent \textbf{Feedback Mechanism: } Carreira \etal \cite{CarreiraMalikCVPR2016} augments input space based on a corrective signal output manifold, improving on human pose estimation and generalising to instance segmentation tasks \cite{LiMalikCVPR2016}. While Wei \etal \cite{wei2016convolutional} used a ConvNet followed by a similar module having a larger receptive field, Newell \etal \cite{NewellHourGlassECCV} developed an hour-glass network design, stacked together for merging information across all input scales. Zamir \etal \cite{ZamirMalikCVPR2017} proposed a novel network architecture aligned with a feedback notion, functionally akin to ResNet architecture. Others include instance segmentation \cite{LiMalikCVPR2016}, few-shot learning \cite{zhang2019canet}, object detection \cite{chen2018iterative}, super resolution \cite{HarisUkitaCVPR2018} and image generation \cite{ShamaManorAFL2018, HuhZhangCVPR2019}. However, we here introduce cross-modal iterative feedback for text recognition with a conditional variational autoencoder that enables modelling of a prior knowledge of linguistically correct character-sequences.

\vspace{0.08cm}
\noindent \textbf{Error Correction:}
 Earlier efforts to utilise this idea of refining character-sequence prediction can be found in automatic speech recognition (ASR) community. In works such as Rozovskaya \etal \cite{RozovskayaCorrection} and Hans \etal \cite{HanCorrection} preposition errors were detected using classifiers. Grammatical error correction was approached using statistical machine translation methods by Ng \etal \cite{ng-etal-2014-conll}. Recently,the idea of using recurrent encoder-decoder architecture in conjunction with attention mechanism was presented by Xie \etal \cite{XieNG2016} to address complex orthographic errors. Later this idea was adopted by Guo \etal \cite{GuoWeiss2019} for tasks related to ASR.

\begin{figure*}[t]
	\begin{center}
		\includegraphics[width=1.0\linewidth]{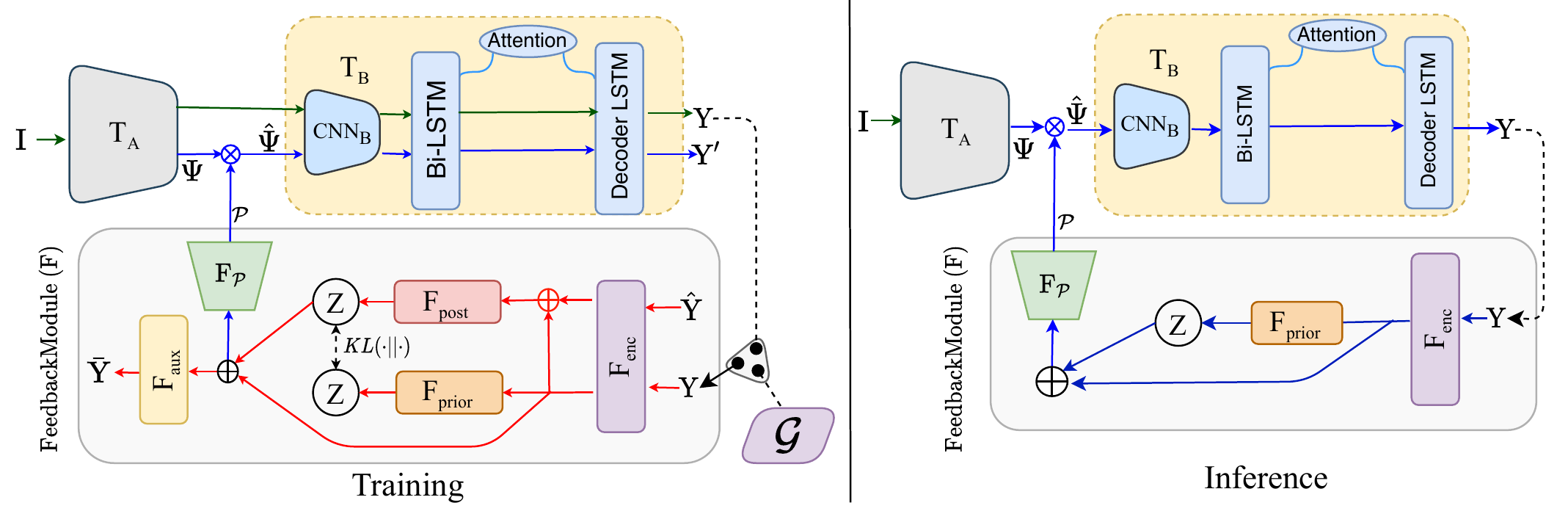}
	\end{center}
\vspace{-0.55cm}
	\caption{\textbf{Training of the network} consists of 2 steps: (i) Predicting Y (green arrow); (ii) Refining Y to $Y'$ using feedback mechanism. Feedback module excluding $F_{\mathcal{P}}$ can be trained using \textit{additional} text-only data $\mathcal{G}$ (red). However, when trained using prediction $Y$, the entire path is followed (red+blue). \textbf{Inference in the model} consists of predicting $\mathrm{Y}^t$ using any arbitrary number of correction steps.}
	\label{fig:short}
	\vspace{-0.5cm}
\end{figure*}

\section{Methodology}

Instead of generating results in a single feed forward pass \cite{ShiBaiPAMI2019, moran, yang2019symmetry, DomainAdaptationCVPR2019}, we propose an iterative approach towards text recognition. As our work solely contributes to the text recognition part, we have refrained from detailing on the initial rectification network. We have used an off-the-shelf rectification network based on Spatial Transformer Network \cite{jaderberg2015spatial}  and Thin Plate Splines \cite{baek2019wrong} from Shi \etal \cite{ShiBaiPAMI2019}.

\subsection{Text Recognition Module}
 The rectified image received from the rectification network \cite{ShiBaiPAMI2019} is fed into this text recognition network  $\mathbf{T}$ aiming to produce a character sequence $\mathrm{Y} = \{ y_{1},  y_{2}, ...,  y_{K}\}$, where $K$ denotes the variable length of text. Given an image $\mathrm{I} \in \mathbb{R}^{H\times W\times C}$, the convolutional feature extractor tries to learn rich visual information and produce a feature map of size $\mathbb{R}^{H'\times W'\times D}$, where $H'$, $W'$, and $D$ are the height, width and number of channels in the output feature map respectively. That output is reshaped into a sequence of feature vectors $\mathrm{B} = [b_{1}, b_{2}, ..., b_{L}]$, where $L=H' \times W'$, and $b_{i} \in \mathbb{R}^{D}$. Every D dimensional feature vector $b_{i}$ encodes a particular local image region based on its receptive fields. Thereafter a bidirectional-LSTM is employed to capture the long range dependency in both the directions, thus alleviating the constraints of limited receptive fields. It outputs an updated feature sequence of same length, denoted by $\mathrm{H} \in [h_{1}, h_{2}, ..., h_{L}]$. Following this $y_{k}$ is decoded based on three factors viz; bidirectional-encoder output $\mathrm{H}$, the previous internal state  $s_{k-1}$, and the character $y_{k-1}$ predicted in the last step. At time step $k$, a recurrent decoder network generates an output vector $o_{k}$ and a new state vector $s_{k}$, defined as: $\mathrm{(o_{k}, s_{k})  = RNN(s_{k-1}, [g_{k}, E(y_{k-1})] )}$ where $g_{k}$ is the glimpse vector \cite{ShiBaiPAMI2019} that encodes the information from specific relevant parts of the encoded feature $\mathrm{H}$ to predict  $y_{k}$; $\mathrm{E}$ is an embedding layer, and $[\cdot]$ signifies a concatenation operation. Here, $g_{k}$ is computed as:  $g_{k} =  \sum_{i=0}^{L} \left (\frac{\exp(a_{k, i})}{ \sum_{j=0}^{L} \exp(a_{k,j})}  \right ) h_{i}$ and, attention score $\mathrm{a_{k,i} = v^{\top}tanh(W_{s}s_{k-1} + W_{h}h_{i} + b_{a})}$, where $\mathrm{v}, \mathrm{W_{s}}, \mathrm{W_{h}}, \mathrm{b_{a}}$ are the learnable parameters. Finally the current step character $y_{k}$  is predicted by: $\mathrm{p(y_{k}) = softmax(W_{o}o_{k} + b_{o})}$ where $W_{o}$ and  $b_{o}$ are trainable parameters. 
 
\subsection{Cross-Modal Variational Feedback}\label{cvae}
\noindent \textbf{Overview:} Unlike previous attempts of devising mostly deterministic feedback modules \cite{HuhZhangCVPR2019, ShamaManorAFL2018, ZamirMalikCVPR2017}, we propose a Cross-Modal Variational Feedback network instilling benefits of variational autoencoder (VAE) \cite{kingma2013auto}, a powerful class of probabilistic models. Let us consider a text recognition network  $\mathbf{T}$ that has been dissected into two parts, namely $\mathbf{T_{A}}$ and $\mathbf{T_{B}}$ where the particular position of dissection is obtained empirically as described in section \ref{ablation}. The prior knowledge from the resultant discrete character space is modelled using a \emph{Feedback Network} $\mathbf{F}$ by predicting affine transformation parameters. Such parameters modulate the output activation map of $\mathbf{T_{A}}$, as $\mathrm{\Psi = T_{A}(I)} \in \mathbb{R}^{\hat{H} \times \hat{W} \times \hat{T}}$, via a designed feedback conditioning layer $\mathrm{\Phi}$. In other words, for iteration step \textit{t}, the feedback network takes the  preceding iteration's output $\mathrm{Y^{t-1}}$ as its input and predicts the transformation parameters $\mathcal{P}^{t}$ as its output. As a result it learns the mapping: $\mathrm{\mathbf{F}:  Y \rightarrow \mathcal{P}}$, such that feedback conditioning layer $\mathrm{\Phi}$ modulates $\mathrm{\Psi}$ based on $\mathcal{P}^{t}$. This can be depicted as $\mathrm{\hat{\Psi^{t}} = \Phi(\Psi; \mathcal{P}^{t})}$ where $\mathrm{\hat{\Psi^{t}}}$ being fed to $\mathbf{T_{B}}$ predicts $\mathrm{Y^{t}}$ with higher precision. 

\vspace{0.1cm}
\noindent \textbf{Feedback Conditioning Layer:} 
The primary objective of this layer is the propagation of prior knowledge from previous feed-forward pass prediction to $\mathbf{T}$. This action couples the rich visual information extracted by $\mathbf{T_{A}}$, with the prior feedback signal coming from earlier prediction. A noteworthy mention in this context would be the work by Perez \etal \cite{perez2018film}, where a general purpose conditioning layer FiLM has been designed based on simple feature-wise affine transformation operation for visual reasoning. Works involving visual-question-answering \cite{perez2018film}, image style transfer \cite{huang2017arbitrary} and semantic image synthesis \cite{park2019semantic} have been seen to endorse a similar idea as well. Based on some prior or conditional information, the intermediate activation map $\mathrm{\Psi}$ could be modulated as: $\mathrm{\hat{\Psi}  = \Psi \odot \gamma + \omega}$, where $\mathrm{\gamma}$, $\mathrm{\omega}$ are \textit{global} affine transformation parameters. They both have dimensions $\mathbb{R}^{\hat{T}}$, and are usually predicted by a network receiving the conditional information as input. In lieu of transforming each channel globally, we allow local transformations \cite{park2019semantic} that attune each spatial position of the activation map. Following earlier mathematical notations and assumptions, the feedback conditioning layer can be formulated as: $\mathrm{\hat{\Psi} = \Psi\odot \Gamma + \Omega}$ where $\{ \Gamma, \Omega\} \triangleq \mathcal{P}$ are the \textit{local} transformation parameters predicted by the feedback network having dimension of $\mathbb{R}^{\hat{H} \times \hat{W} \times \hat{T}}$, similar to $\mathrm{\Psi}$. This layer couples the prediction $Y$ with visual feature $\mathrm{\Psi}$ by scaling, shifting and ReLU influenced selective thresholding, alternatively. Compared to global tuning \cite{perez2018film}, such local harmonisation of features ensure a better fine-grained control over the activations at each layer.

\vspace{0.1 cm}

\noindent \textbf{Feedback Network:} Our goal is to model a conditional distribution $p(\mathcal{P} | Y)$ from a discrete predicted character space $Y$ to the transformation parameter space $\mathcal{P}$ of feedback conditioning layer via a feedback network based on CVAE \cite{sohn2015learning}.  Sohn \etal \cite{sohn2015learning} has shown that the variational lower bound of this conditional distribution can be written as:

 \vspace{-0.7 cm}
\begin{multline}
\label{obj1}
\mathrm{\mathcal{\widetilde{L}}(Y, \mathcal{P}; \theta, \psi) = -KL(q_{\psi}(z|\mathcal{P}, Y) || p_{\theta}(z|Y)) +} \\ \mathrm{\mathbb{E}_{q_{\psi}}[log \; p(\mathcal{P}|z,Y)] \leq log  \; p( \mathcal{P}| Y)}
 \end{multline}
\vspace{-0.55cm}
 
\noindent where $z$ is a latent variable assumed to follow a multivariate Gaussian distribution with a diagonal covariance matrix. Due to intractability of true posterior, we approximate the posterior distribution of $z$ through a recognition neural network $q_{\psi}(z|\mathcal{P}, Y)$. The prior distribution of $z$ given $\mathrm{Y}$ is modelled through a prior network $p_{\theta}(z|Y)$. For more details about CVAE, we refer to \cite{sohn2015learning}.

\indent \textbf{(i)}
Ideally speaking, $\mathcal{P}$ is supposed to adjust the activation map $\mathrm{\Psi}$ in a way that predicts $\hat{Y}$ when passed through a fixed $\mathbf{T_{B}}$. In other words, for time-step $t$, given $\mathcal{P}^{t}$ predicted by the feedback network with a conditioning on $Y^{t-1}$, the modulated feature map $\noindent{\hat{\Psi^{t}}}$ is fed into $\mathbf{T_{B}}$ obtains the output of next iteration $\mathrm{Y'}$. Therefore, $\mathrm{Y' = \mathbf{T_{B}}(\hat{\Psi^{t}})}$, where $\mathrm{\hat{\Psi^{t}} = \Phi(\Psi; \mathcal{P}^{t})}$. Thus, minimizing the cross-entropy loss between $Y'$ and $\hat{Y}$ is equivalent to maximizing the likelihood of $\mathcal{P}$. 
 
\indent \textbf{(ii)} We assume that posterior of $z$ depends on the actual ground-truth label $\hat{Y}$ instead of $\mathcal{P}$, so $q_{\psi}(z|\mathcal{P}, Y) \approx q_{\psi}(z|\hat{Y}, Y)$ which effectively makes the latent space $z$ aware of exact ground-truth character sequence $\hat{Y}$.

\indent \textbf{(iii)} Inspired by the auxiliary task approach towards improving primary task objective \cite{DiscourseLevelDiversity2017}, we aim to decode the ground truth character sequence $\hat{Y}$ directly from $z$ by using an auxiliary character sequence decoder. Firstly, this approach tackles the problem of vanishing latent variable \cite{bowman2015generating} and provides better gradient to regularize the learning of feedback module. Secondly, we discover a choice of training our feedback module (refer Figure \ref{fig:short}) via relativistic information of \textit{prediction:ground-truth pairs} (say $\mathcal{G}$) generated from other state-of-the-arts, instead of depending on $\mathbf{T}$ predicted $Y$ alone. This helps the module in learning an associative relation between candidate correct words and closely related erroneous instances. Loosely speaking `$hello$' might be predicted as `$nello$' or `$bello$' due to a partial structural similarity of `$h$' with `$n$' or `$b$'. Learning this sense from \emph{error distribution} imparts the model a semantic knowledge of required character association to form a valid word. Additionally, using text-only data $\mathcal{G}$ alleviates the issue of limited availability of image-paired datasets during training. 

Hence, we adapt Eqn.~\ref{obj1} for our feedback network to generate a prior knowledge $\mathcal{P}^{*}$, instead of transformation parameters only. This prior knowledge has two components. One encapsulates relationship between $Y$ and $\hat{Y}$, while the other generates $\mathcal{P}$ and injects $Y$ into $\mathbf{T}$ for next prediction.
Therefore adopting the aforementioned variational lower bound expression (Eqn. \ref{obj1}), and assuming conditional independence of those two knowledge components (given $z$ and $Y$) denoted by $\mathrm{p(\mathcal{P}^{*}|z, Y) = p(\mathcal{P}|z, Y)p(\hat{Y}|z, Y)}$, we get a modified lower bound as:

 \vspace{-0.6cm}

\begin{multline}\label{final_lower_bound}
\mathrm{\mathcal{\widetilde{L'}}(Y, \mathcal{P}^{*}; \theta, \psi) = -KL(q_{\psi}(z|\hat{Y}, Y) || p_{\theta}(z|Y)) +} \\ \mathrm{\mathbb{E}_{q_{\psi}}[log \; p(\mathcal{P}|z,Y)] +  \mathbb{E}_{q_{\psi}}[log \; p(\hat{Y}|z,Y)]}
\end{multline}
 \vspace{-0.6cm}

\noindent \textbf{Network Components:} Ignoring the time-step notation, let us consider any word image $I$ having ground-truth $\hat{Y}$, that undergoes the first forward-pass of $\mathbf{T}$ predicting $Y$ and $Y'$ in current and successive post-feedback iterations respectively. During training, we obtain an embedding representation of both $\hat{Y}$ and $Y$ through a shared encoder network $\mathbf{F}_{enc}$: $\hat{Y}_{enc} = \mathbf{F}_{enc}(\hat{Y})$ and $Y_{enc} = \mathbf{F}_{enc}(Y)$. Moving on we have two independent branches: A posterior network $\mathbf{F}_{post}$ to estimate parameters of posterior distribution, and a prior network $\mathbf{F}_{prior}$ to do the same for prior distribution. Following this, we get: $\mathrm{\mu_{post}, \sigma_{post} = \mathbf{F}_{post}([Y_{enc},\hat{Y}_{enc}])}$ and $\mathrm{\mu_{prior}, \sigma_{prior} = \mathbf{F}_{prior}(Y_{enc})}$. A latent variable $z$ is sampled from the posterior (or prior during testing) distribution and merged with $Y_{enc}$ before feeding it to the auxiliary ground-truth character sequence decoder $\mathbf{F}_{aux}$ and transformation parameter prediction sub-module $\mathbf{F}_{\mathcal{P}}$. Therefore we have, $\mathcal{P} = \mathbf{F}_{\mathcal{P}}([Y_{enc}, z])$, and $\bar{Y} = \mathbf{F}_{aux}([Y_{enc}, z])$ where $\mathcal{P}$ encapsulates both $\Gamma$ and $\Omega$, and $\mathbf{\bar{Y}}$ is output from the auxiliary decoder ${F}_{aux}$. Thereafter, we evaluate the prediction of next iteration as: $\mathrm{Y' = \mathbf{T_{B}}(\hat{\Psi})}$, where $\mathrm{\hat{\Psi} = \Phi(\Psi; \mathcal{P}})$. Please refer to Figure \ref{fig:short} for clarity.

\keypoint{Learning Objectives:}  Our baseline text recognition network $\mathbf{T}$ (parameters $\mathrm{\mathbf{\theta_{T}}}$) is trained using Cross-Entropy (ce) loss summed over the ground-truth output sequence $\mathrm{\hat{Y}}$
\vspace{-0.35cm}
\begin{equation}\label{text_3}
 \mathrm{L^{\mathbf{\theta_{T}}} = \mathrm{L_{ce}(Y, \hat{Y}) = -  \sum_{k=1}^{K} \hat{y_{k}} logP (y_{k}  | I, \hat{y_{k-1}}})}
\vspace{-0.3cm}
\end{equation}
 
Feedback module $\mathbf{F}$ is to be trained from two input sources: (a) by using the prediction $Y$ obtained from $\mathbf{T}$. In cases where the forward-pass prediction $Y$ is accurate the very first time, a diverged prediction value in the next iteration is highly undesirable. 
Hence, along with the lower bound (Eqn. \ref{final_lower_bound}), we impose a monotonically decreasing constraint $L_{c}$. This enforces the loss value (Eqn. \ref{text_3}) related to the current iteration $\{Y', \hat{Y} \}$ to be lesser than its previous $\{Y, \hat{Y}\}$, thus converging predictions to a higher precision. Hence, we optimize all the parameters $\mathrm{\mathbf{\theta_{F}}}$ of feedback module using: 

\vspace{-0.7cm}
\begin{multline} \label{feedback_loss} 
 \mathrm{L^{\mathbf{\theta_{F}}} = \lambda_{1}L_{ce}(Y', \hat{Y}) + \lambda_{2}L_{ce}(\bar{Y}, \hat{Y}) + \lambda_{3}L_{KL} + \lambda_{4}L_{c}} \\
     \mathrm{where, L_{c} = max(0, L_{ce}(Y', \hat{Y}) - L_{ce}(Y, \hat{Y}))}
\end{multline}
  \vspace{-0.7cm}

\noindent (b) By using pre-collected text-only data $\mathcal{G}$ generated from existing text-recognition methods \cite{moran, AONRecogCVPR2018, baek2019wrong, ShiBaiPAMI2019}. Doing so develops a semantic sense of associative relation between candidate correct words and erroneous alternatives. Keeping $F_{\mathcal{P}}$ fixed, rest of the feedback module (parameters denoted by $\mathbf{\theta_{F'}}$) is optimised solely using auxiliary decoder reconstruction loss and KL divergence loss as:  

\vspace{-0.3cm} 
\begin{equation}\label{feedback_aux} 
 \mathrm{L^{\mathbf{\theta_{F'}}}= \lambda_{2}L_{ce}(\bar{Y}, \hat{Y}) + \lambda_{3}L_{KL}}    
\end{equation} \label{feedback_loss2}
\vspace{-0.5cm} 

\noindent During testing we exclude both $\mathbf{F}_{aux}$ and $\mathbf{F}_{post}$, where iterative prediction is done using the transformation parameter predicted by $\mathbf{F}_{\mathcal{P}}$ (see Figure \ref{fig:short}).  
 
\keypoint{Comparative Discussions:} While state-of-the-art text recognition frameworks model a conditional distribution $\mathrm{p(Y|I)}$, our modelling objective is $\mathrm{p(Y^{t}|Y^{t-1},I)}$. This could be decoupled into two marginal distributions related via a transformation parameter space $\mathcal{P}^{t}$ (more strictly prior knowledge $\mathcal{P}^{*t}$). Assuming the obvious conditional independence, we reformulate it as: $\mathrm{p(Y^{t}| Y^{t-1},I) = p(Y^{t}|I,\mathcal{P}^t) \times  p(\mathcal{P}^t|Y^{t-1})}$, where $\mathrm{p(Y^{t}|I,\mathcal{P}^t)}$ is our modified feed forward text recognition model and  $\mathrm{p(\mathcal{P}^t|Y^{t-1})}$ resembles our feedback network.
 
\setlength{\textfloatsep}{4pt}
\begin{algorithm}[!hbt]
\sloppy
	\caption{Training algorithm of the proposed framework}
	\begin{algorithmic}[1] 
        \State \textbf{Input}: Image and ground-truth pairs $\mathcal{D}$;  
             prediction and ground-truth pairs $\mathcal{G}$. 
        \label{algo}
        \State \textbf{Initialise hyper params}: $\alpha_{1}$, $\alpha_{2}$ be learning rate for  $\mathbf{T}$, $\mathbf{F}$ respectively. 
        \State \textbf{Initialise model params}: $\theta_{T}$, $\theta_{F}$ ($\theta_{F'} \subset \theta_{F}$  excluding $F_{\mathcal{P}}$)
        \While {not done training}
            
            \State Sample a mini-batch $\mathcal{D}_{i}$  from $\mathcal{D}$ 

             \State \hskip2em Update $\theta_{T} := \theta_{T} - \alpha_{1} \nabla_{\theta_{T}}(L^\mathbf{\theta_{T}})$   \Comment{Eqn.~\ref{text_3}}
            
             \State \hskip2em Update $\theta_{F} := \theta_{F} - \alpha_{2} \nabla_{\theta_{F}}(L^\mathbf{\theta_{F}})$ \Comment{Eqn.~\ref{feedback_loss}}  

            \State Sample a mini-batch $\mathcal{G}_{i}$  from $\mathcal{G}$ 

             \State \hskip2em Update $\theta_{F'} := \theta_{F'} - \alpha_{2} \nabla_{\theta_{F'}}(L^\mathbf{\theta_{F'}})$   \Comment{Eqn.~\ref{feedback_aux}}
        \EndWhile
     \State \textbf{Output}:  $\theta_{T}$, $\theta_{F}$    
	\end{algorithmic} 
\vspace{-0.15cm}	
\end{algorithm}

\vspace{-0.5cm} 
\section{Experiments}\label{sec:experiments}

\begin{table*}[h]

\centering
\caption{Comparison of unconstrained WRA for \textbf{novel words not encountered during training}. \textit{$t=0$ signifies no feedback.}}
\resizebox{1.6\columnwidth}{!}{
\begin{tabular}{c|c|c|c|c|c|c|c|c}
\hline
\textbf{Methods} & \textbf{IIIT5K} & \textbf{SVT} & \textbf{IC13} & \textbf{IC15} & \textbf{SVTP} & \textbf{CUTE80} & \textbf{IAM} & \textbf{RIMES} \\
\hline

\multirow{6}{*}{$\left\{\begin{array}{l}
\text{Shi \etal \cite{ShiBaiPAMI2019} (t=0) No-Feedback} \\
\text{Baseline Seq-SCM} \\
\text{Baseline Deterministic-Feedback} \\
\text{Shi \etal \cite{ShiBaiPAMI2019} \textbf{+ CVAE-Feedback} (t=1)} \\
\text{Shi \etal \cite{ShiBaiPAMI2019} \textbf{+ CVAE-Feedback} (t=2)} \\
\text{Shi \etal \cite{ShiBaiPAMI2019} \textbf{+ CVAE-Feedback} (t=3)} \\
\end{array}\right.$}

& 84.3 & 84.2 & 82.6 & 65.7 & 74.4 & 61.6 & 54.3 & 59.7 \\
& 85.6 & 84.1 & 83.7 & 65.5 & 75.8 & 63.4 & 57.6 & 63.7 \\
& 87.9 & 86.8 & 85.9 & 70.4 & 78.6 & 64.7 & 59.9 & 69.7 \\
& 90.6 & 88.7 & 89.3 & 72.2 & 79.6 & 65.1 & 64.5 & 70.4 \\
& \textbf{90.8} & \textbf{88.9} & \textbf{89.4} & \textbf{72.6} & \textbf{79.6} & \textbf{66.1} & \textbf{64.8} & \textbf{70.5} \\
& 90.7 & 88.8 & 89.4 & 72.5 & 79.6 & 65.8 & 64.6 & 70.3 \\
\hline
\textbf{Relative Gain} (t=0 vs t=2) & 6.5$\uparrow$ & 4.7$\uparrow$ & 6.8$\uparrow$ & 6.9$\uparrow$ & 5.2$\uparrow$ & 4.5$\uparrow$ & 10.5$\uparrow$ & 10.8 $\uparrow$ \\

\hline \hline
\vspace{0.9pt}

\multirow{2}{*}{$\left\{\begin{array}{l}
\text{Show, Attend and Read \cite{li2019show} (t=0) No-Feedback} \\
\text{Show, Attend and Read \cite{li2019show} \textbf{+ CVAE-Feedback} (t=2)} \\
\end{array}\right.$}

& 85.8 & 86.5 & 84.7 & 68.4 & 82.2 & 71.8 & 57.9 & 62.8 \\
& \red{\textbf{91.5}} & \textbf{90.5} & \red{\textbf{91.2}} & \textbf{74.8} & \textbf{87.1} & \red{\textbf{75.0}} & \textbf{68.0} & \textbf{73.0} \\
\Xhline{1pt} 
\textbf{Relative Gain} (t=0 vs t=2) & 5.7$\uparrow$ & 4.0$\uparrow$ & 6.5$\uparrow$ & 6.4$\uparrow$ & 4.9$\uparrow$ & 3.2$\uparrow$ & 10.1$\uparrow$ & 10.2 $\uparrow$ \\
\hline \hline

\vspace{0.9pt}

\multirow{2}{*}{$\left\{\begin{array}{l}
\text{SCATTER \cite{litman2020scatter} (t=0) No-Feedback} \\
\text{SCATTER \cite{litman2020scatter} \textbf{+ CVAE-Feedback} (t=2) } \\
\end{array}\right.$}

& 84.7 & 86.9 & 84.3 & 71.8 & 82.6 & 69.3 & 59.0 & 62.9 \\
& \textbf{91.1} & \red{\textbf{90.9}} & \textbf{90.0} & \red{\textbf{77.7}} & \red{\textbf{87.3}} & \textbf{72.7} & \red{\textbf{68.7}} & \red{\textbf{73.1}} \\
\Xhline{1pt} 
\textbf{Relative Gain} (t=0 vs t=2) & 6.4$\uparrow$ & 4.0$\uparrow$ & 6.6$\uparrow$ & 5.9$\uparrow$ & 4.7$\uparrow$ & 3.4$\uparrow$ & 9.7$\uparrow$ & 10.2 $\uparrow$ \\
\Xhline{1pt} 
\end{tabular}
}
\label{table_big1}
\vspace{-0.35cm}
\end{table*}

\vspace{-0.1cm}
\keypoint{Datasets:} We train our method using an approach similar to \cite{ESIR2019,yang2019symmetry, baek2019wrong, AONRecogCVPR2018, ShiBaiPAMI2019, moran}, on synthetic datasets such as Synth90k \cite{jaderberg2014synthetic} and SynthText \cite{SynthText} that hold 8 and 6 million images respectively. \cut{That model is evaluated on datasets (without fine-tuning) holding real images like} Evaluation is done on: \noindent{\textbf{IIIT5K-Words}}, \noindent{\textbf{Street View Text (SVT)}}, \noindent{\textbf{SVT-Perspective (SVT-P)}}, \noindent{\textbf{ICDAR 2013 (IC13)}}, \noindent{\textbf{ICDAR 2015 (IC15)}}, and \noindent{\textbf{CUTE80}}. IIIT5K-Words \cite{IIIT5K-Words} presents randomly picked 3000 cropped word images. Street View Text \cite{WangICCV2011} contains 647 images, mostly being blurred, noisy or having low resolution. SVT-Perspective \cite{SVT-P} offers 645 samples from side-view angle snapshots having perspective distortion. ICDAR2013 \cite{ICDAR2013} presents 1015 words while ICDAR2015 \cite{ICDAR2015} has 2077 images, 200 of which are irregular. CUTE80 \cite{CUTE80} distinguishes itself by presenting 288 cropped high quality curved text images. Handwritten Text Recognition (HTR) and Scene-Text Recognition (STR) both share a common objective in terms of recognition, which is usually handled by similar network architecture. Hence, we validate our results on two different public HTR datasets. The evaluation setup described in \cite{bhunia2019handwriting} is employed on two large standard datasets viz, \noindent{\textbf{IAM}} \cite{IamDataset} containing 1,15,320 words and \noindent{\textbf{RIMES}} having 66,982 words. For IAM we use the same partition for training, validation and testing as provided. For RIMES, we follow the partition released by ICDAR 2011 competition.

\subsection{Implementation Details}
\vspace{-0.1cm}
\noindent \textbf{Network Design:} Keeping text rectification and recognition networks similar to Shi \etal \cite{ShiBaiPAMI2019}, we implement our framework using open sourced codes \cite{baek2019wrong, moran} in PyTorch \cite{paszke2017automatic}. A bi-directional LSTM of hidden size 256 is used in designing $F_{enc}$ of our feedback module which accepts a one layer MLP embedded 128-dimensional representation of discrete character sequences. For both posterior $F_{post}$ and prior $F_{prior}$ networks, we use 2 layer MLPs with \textit{tanh} non-linearity. The latent variable $z$ has a size of 256. The auxiliary decoder $F_{aux}$ is a one layer LSTM decoder whose initial hidden state is initialized by applying a FC-layer on the concatenated representation of $Y_{enc}$ and sampled $z$. The parameter prediction network $F_{\mathcal{P}}$ is a convolutional decoder network inspired from \cite{zheng2019pluralistic}. The first layer is a fully-connected layer implemented through a 1x1 convolution that maps to a tensor of size equal to the last CNN layer of $\mathbf{T}$ via a reshaping operation. Thereafter we introduce a sequence of residual decoder blocks which upsamples feature maps to higher spatial dimensions in the reversed order of down-sampling followed by $\mathbf{T}$. In other words, if a ResNet encoder block in $\mathbf{T}$ halves the height of feature map, the corresponding ResNet decoder block \cite{zheng2019pluralistic} in $\mathbf{F}_{\mathcal{P}}$ will double it. This strategy essentially formulates intermediate feature maps of a decoder ResNet block ($\mathbf{F}_\mathcal{P}$), whose spatial size is similar to the corresponding layer in the encoder ResNet block ($\mathbf{T}$), thus predicting required affine transformation parameters of similar dimensions as well. Since we need to predict both $\Gamma$ and $\Omega$, we double the number of convolutional filters in the last layer of $\mathbf{F}_\mathcal{P}$ and split the output channel-wise to obtain $\Gamma$ and $\Omega$ individually. 

\keypoint{Training Details:} It has been observed in practice, that warming up individual components in the initial phase, followed by a joint training operation provides better stability than training the entire framework at one go. Text recognition network (\textbf{T}), along with rectification network, is trained using ADADELTA optimizer having a learning rate of 1.0 and a batch size of 64. Meanwhile Feedback module $\mathbf{F}$ is trained via Adam optimizer with a learning rate 0.001, and gradient clipping at 5. In the warm-up phase, firstly, text recognition and rectification networks are trained from a union of MJSynth and SynthText datasets, for 600K iterations. The rectification network is frozen thereafter. Then, to capture a linguistic prior from text-data, the feedback module $\mathbf{F}$ is trained independently from $\mathcal{G}$ for 300K iterations, thereby ignoring the $F_{\mathcal{P}}$ part (Eqn. \ref{feedback_aux}). Finally, keeping $\mathbf{T}$ fixed, we train the complete feedback module using $Y$ (Eqn.~\ref{feedback_loss}), with the same training specifications for 300K iterations. Now, for the joint training (see algorithm \ref{algo}), after updating $\mathbf{T}$, $\mathbf{F}$ is updated with the prediction $Y$ from $\mathbf{T}$, as input. Thereafter, keeping $F_{\mathcal{P}}$ fixed, $\mathbf{F}$ is updated using $\mathcal{G}$. During this training, we reduce the learning rate of $\mathbf{T}$ to 0.01 which continues for 600K iterations. Owing to corresponding data sizes in HTR, iterations for warm-up phases are 100K, 50K and 50K respectively, while for joint training it is 100K. For both STR and HTR, we resize the image to 32x100 and train our model in a 11 GB Nvidia RTX 2080-Ti GPU. $\lambda_{1}$, $\lambda_{2}$, $\lambda_{3}$ and $\lambda_{4}$ has been assigned the values of 1, 1, 50 and 0.5 respectively for the same purpose. In order to generate the text-only data $\mathcal{G}$, we follow an approach similar to \cite{zheng2019distraction}. The entire \textit{training} dataset is divided into train-validation split in a cross validation setting, where top ten beam search decoded hypotheses are collected from different state-of-the-art (SOTA) models \cite{moran, AONRecogCVPR2018, baek2019wrong, ShiBaiPAMI2019}, to harness (using open-source code) the error information from closest possible erroneous alternatives. Such a collection may consist of correctly or incorrectly predicted words. This requires the feedback module to learn a language prior for fixing a wrongly predicted word, as well as for a one-to-one character-sequence mapping for correctly predicted words.

\begin{figure*}[] 
	\begin{center}
	\label{fig3}
		\includegraphics[width=1.\linewidth]{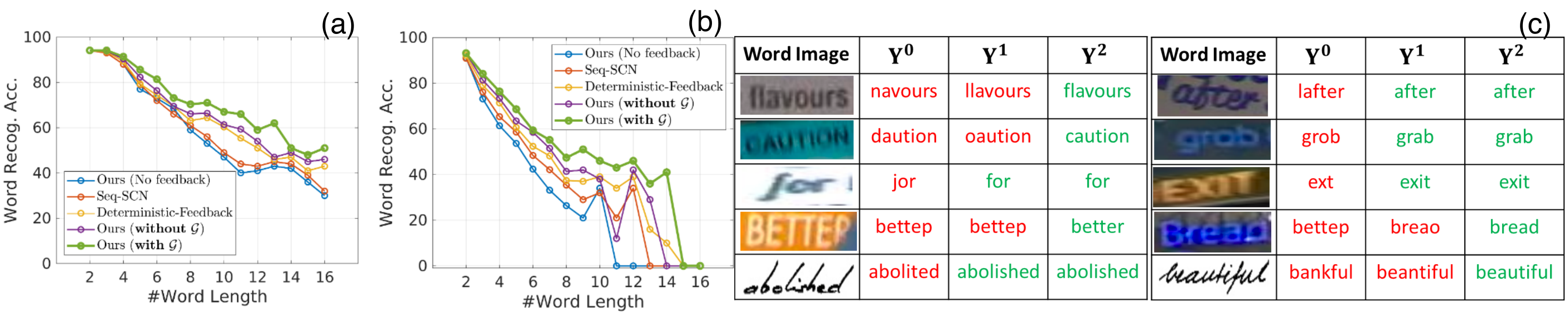}
	\end{center}
	\vspace{-0.2in}
	\caption{Unconstrained WRA on IAM dataset with varying \emph{word length} in (a) CS, (b) DS setups. (c) Few samples which defeat single feed forward pass mechanisms, but are acceptably recognised via our iterative framework in CS setup.
	}
	\label{fig:Graph1_Full}
\vspace{-0.5cm}
\end{figure*}

\vspace{-0.05cm}
\subsection{Performance Analysis}
\vspace{-0.1cm}
\keypoint{Baselines:}
Aligned to our iterative approach we explore two alternatives as baselines.  \textbf{Seq-SCN}: Inspired from Automatic Speech Recognition community \cite{XieNG2016}, a naive baseline could be designed where we train an independent Spelling Correction Module (SCM) based on sequence-to-sequence architecture from text-only training data \cite{GuoWeiss2019} consisting of paired model-hypothesis and corresponding ground-truth. \textbf{Deterministic-Feedback}: Here, we simply replace the CVAE based feedback module by a deterministic encoder(bi-LSTM)-decoder (parameter prediction network) architecture along with Shi \etal \cite{ShiBaiPAMI2019}.  

\keypoint{New Evaluation Setup:} Here we design a new disjoint train-test split (\textbf{DS}) in addition to conventional train-test split (\textbf{CS}). While training we remove all words from MJSynth and Synth90K whose truth-pairs appear in any of the mentioned STR testing datasets. Additionally we ensure, $\mathcal{G}$ \textit{does not contain} any information from the testing set. Due to size constraint in HTR datasets, we split it such that all word-image pairs corresponding to one particular ground-truth character sequence must collectively fall in either one of training or testing set, thus ensuring disjointedness. This evaluation protocol grades our model's recognition performance on word images whose ground-truth character sequences never appeared in the training dataset. We use it to verify generalising capability of our model for unseen or rarely seen words. Furthermore, superiority of our model in this scenario confirms a fair result on datasets having \emph{less unique words}. As large datasets are rarely available for text recognition (specifically handwriting) apart from English, it is ideal to use a dataset created by collecting multiple instances of a small set of unique words which thus need to be annotated just once per unique word.
To do so any model needs to learn character-specific fine-grained details from a small available set of unique words, and generalise onto other unseen character sequences.  {Please note that we re-train SOTA models \cite{moran, AONRecogCVPR2018, baek2019wrong, ShiBaiPAMI2019},  used to collect the error distribution,  by ensuring no words from evaluation set appear in their training (ensuring no ``leak" of information), while collecting error distribution for DS setup.}

\noindent \textbf{Result Analysis:} Along with the designed baselines, we incorporate our iterative design on top of three popular state-of-the-art (SOTA) feed-forward text recognition frameworks -- a) \textbf{Shi \etal} \cite{ShiBaiPAMI2019}, b) \textbf{Show, Attend and Read} \cite{li2019show} c) \textbf{SCATTER} \cite{litman2020scatter}. \textbf{Show, Attend and Read} \cite{li2019show}  extends \cite{ShiBaiPAMI2019} by including 2D attention and SCATTER \cite{litman2020scatter} couples multiple BLSTM layers for richer context modelling. We follow similar training protocol in \cite{ShiBaiPAMI2019, li2019show, litman2020scatter} respectively. Nevertheless, ours is a \textbf{meta-framework} and could be added on top of most SOTA frameworks. Table \ref{table_big1}  (highest scores are in \red{red}) depicts unconstrained word recognition accuracy (WRA) on unseen words (DS setup). [i] \textbf{Comparison with Baselines:}  \textit{Seq-SCN} performs inferior to our method as the rich visual feature is not harnessed while refining the prediction. At times it fails to copy an already accurate prediction, leading to a lower accuracy in certain datasets. \textit{Deterministic-Feedback} being an iterative framework, performs better than other baselines at $t=2$, however, it lags behind our design since any uncertainty handling potential is absent. [ii] \textbf{Significant improvement under DS setup:}  From  Table \ref{table_big1}, improvement due to our iterative pipeline is quite evident in the DS scenario over three SOTA baselines \cite{ShiBaiPAMI2019, li2019show, litman2020scatter} . Similarly, in handwritten dataset (Table \ref{table_big1}), the performance drop in DS setting is way more severe than its STR counterpart. This signifies that HTR poses a greater challenge due to its free-flow nature of writing. Improvement against Shi \etal \cite{ShiBaiPAMI2019} reaches to 10.5\% and 10.8\%, without lexicon information (unconstrained) in HTR dataset of IAM and RIMES respectively, while for STR dataset of IC15 it secures a 6.9\% rise in DS setup. [iii] \textbf{Additional Observations:} Improvement at $t$=2 w.r.t $t$=0 is shown as relative gain where our method outperforms fairly in CS setup, and largely in DS setup (Table \ref{big_table2}). In fact optimal WRA is seen at iteration $t=2$, which then diminishes. Contrary to a possible impression, that our feedback module might remember the erroneous pairings from $\mathcal{G}$, the improvement on ``unseen" words empirically validates against it. 
Figure \ref{fig:Graph1_Full} shows qualitative results.

\vspace{-0.3cm} 
\begin{table}[hbt!]
\normalsize    
\centering
\caption{Comparison with SOTA results using standard setup \cite{baek2019wrong}.}
\resizebox{1.\columnwidth}{!}{
\begin{tabular}{ccccccc}
\hline
\textbf{Methods} & \textbf{IIIT5K} & \textbf{IC13} & \textbf{IC15} & \textbf{SVTP} & \textbf{CUTE80} & \textbf{IAM} \\ 
\hline
Yang \etal \cite{yang2019symmetry} & 94.4 & 93.9 & 78.7 & 80.8 & 87.5 & - \\
Luo \etal \cite{moran} & 91.2 & 92.4 & 68.8 & 76.1 & 77.4 & 82.1 \\
Cheng \etal \cite{AONRecogCVPR2018} & 87.0 & - & 68.2 & 73.0 & 76.8 & - \\
Zhang \etal \cite{DomainAdaptationCVPR2019} & 83.8 & - & - & - & - & -\\
Baek \etal \cite{baek2019wrong} & 87.9 & 92.3 & 71.8 & 79.2 & 74.0 & - \\
Lyu \etal \cite{2Dattentional} & 94.0 & 92.7 & 76.3 & 82.3 & 86.8 & - \\
Zhan \etal \cite{ESIR2019} & 93.3 & - & 76.9 & 79.6 & 83.3 & - \\
Shi \etal \cite{ShiBai2016} & 81.9 & - & - & - & - & 80.3 \\
Cheng \etal \cite{ChengICCV2017} & 87.4 & 93.3 & 70.6 & - & - & - \\
$\circledast$ Shi \etal \cite{ShiBaiPAMI2019} & 93.4 & 91.8 & 76.1 & 78.5 & 79.5 & - \\
$\circledast$ Li \etal \cite{li2019show} & 95.0 & 94.0 & 78.8 & 86.4 & 89.6 & - \\
$\circledast$ Litman \etal \cite{litman2020scatter} & 93.7 & 93.9 & 82.2 & 86.9 & 87.5 & - \\
\hline
Shi \etal \cite{ShiBaiPAMI2019} (t=0) & 93.2 & 91.6 & 75.9 & 78.2 & 79.3 & 82.3 \\
Li \etal \cite{li2019show} (t=0) & 94.8 & 93.7 & 78.6 & 86.0 & 89.5 & 85.9 \\
Litman \etal \cite{litman2020scatter} (t=0) & 93.6 & 93.8 & 82.0 & 86.5 & 87.0 & 86.0 \\
Baseline Seq-SCM & 93.4 & 91.8 & 75.8 & 78.5 & 79.9 & 82.9 \\
Deterministic-Feed.  & 93.5 & 92.7 & 77.1 & 79.6 & 80.5 & 85.6 \\
\cite{ShiBaiPAMI2019}\textbf{+CVAE-Feed.} (t=2) & \textbf{94.9} & \textbf{93.7} & \textbf{78.8} & \textbf{80.9} & \textbf{82.9} & \textbf{87.5} \\
\cite{li2019show} \textbf{+CVAE-Feed.}(t=2) &  \textbf{\red{96.3}} & \textbf{{95.4}} & \textbf{81.4} & \textbf{88.5} & \textbf{\red{91.0}} & \textbf{89.7} \\
\cite{litman2020scatter} \textbf{+CVAE-Feed.}(t=2) & \textbf{95.2} & \textbf{\red{95.7}} & \textbf{\red{84.6}} & \textbf{\red{88.9}} & \textbf{89.7} & \textbf{\red{90.3}} \\
\Xhline{1pt}
\end{tabular}
}
\label{big_table2}
\end{table}

\vspace{-0.3cm}
\subsection{Further Analysis and Insights} \label{ablation}
\noindent \textbf{Ablation Study:} We have done a thorough ablative study to justify the contribution of every design choice on both IAM (HTR) and IC15 (STR) datasets \texttt{using Shi \etal \cite{ShiBaiPAMI2019} as baseline}. 
\textbf{[i] Significance of auxiliary decoder:} On removing this part along with its respective loss function from Eqn.~\ref{feedback_loss} we see a performance drop of 2.19\%(1.98\%) and 5.01\%(2.94\%) in the CS and DS setting for IAM (IC15) dataset respectively, thus confirming its contribution.
\textbf{[ii] Significance of monotonically decreasing constraint:} Removing it, destabilizes by 5.47\% (3.87\%) in DS setup, for IAM (IC15) datasets respectively, thus confirming its importance. \textbf{[iii] Significance of using error distribution:} Discarding data inclusion from $\mathcal{G}$(\emph{error distribution}), leads to a performance drop of 2.17\% (1.87\%) in conventional train-test split, and a further drop of 4.9\% (2.48\%) in disjoint setup, for IAM (IC15) datasets. Please refer to Table \ref{tab:my-table7} for more analysis. \textbf{[iv] Finding optimal block for feedback:} We evaluated the performance by providing a feedback signal into every ResNet block of backbone feature extractor of $\mathbf{T}$ at a time. Table \ref{table2} shows a complete analysis on both IAM and IC15 datasets in CS and DS setups which shows the optimum result to be obtained by supplying feedback signal to \textit{Block-3} of ResNet convolutional architecture. Moreover, local transformation is seen to outperform the global one. \textbf{[v] Performance with varying text length} It is often observed that any text recognition framework struggles to recognize lengthy words. Owing to its iterative refining approach, along with a modelled linguistic prior, our method shows a considerably higher performance in comparison to no-feedback baseline on increasing character sequence length, as shown in Figure \ref{fig:Graph1_Full}. \textbf{[vi] Computational cost:} Finally, we want to notify that benefit of any iterative pipeline \cite{CarreiraMalikCVPR2016, chen2018iterative, LiMalikCVPR2016} does incur extra computational expenses, be it text rectification \cite{ESIR2019} or in our case text-recognition. A through study on complexity and speed analysis (Intel(R) Xeon(R) W-2123 CPU @ 3.60GHz) in Table \ref{table5} reveals that $T_{B}$ takes most time due to its sequential decoding operation, whereas $F$ adds minimal burden. Please refer to \emph{supplementary} as well.  
\vspace{-0.2cm}  
\begin{table}[!hbt]
\centering
\caption{WRA of using feedback after a specific ResNet Block (abbreviated as `Blk') such as  Block 1 $(16 \times 50 \times 64)$, Block 2 $(8 \times 25 \times 128)$, Block 3 $(4 \times 25 \times 256)$, Block 4 $(2 \times 25 \times 256)$, Block 5 $(1 \times 25 \times 256)$ as described in ASTER \cite{ShiBaiPAMI2019}.}
\label{table2}
\resizebox{1.\columnwidth}{!}{%
\begin{tabular}{c|cc|cc|cc|cc|cc}
\hline
\multirow{3}{*}{Methods} & \multicolumn{10}{c}{Conventional Setup (CS) \& Disjoint Setup (DS)} \\ 
\cline{2-11} 
  & \multicolumn{2}{c|}{Blk\_1} & \multicolumn{2}{c|}{Blk\_2} & \multicolumn{2}{c|}{Blk\_3} & \multicolumn{2}{c|}{Blk\_4} & \multicolumn{2}{c}{Blk\_5} \\
 \cline{2-11}
 & CS & DS & CS & DS & CS & DS & CS & DS & CS & DS \\ \hline
Global(IAM) & 84.7 & 59.9 & 85.5 & 60.8 & 86.7 & 61.9 & 86.2 & 60.8 & 84.5 & 58.5 \\
Local(IAM) & 84.9 & 61.0 & 86.7 & 64.2 & 87.5 & 64.8 & 87.3 & 63.9 & 85.6 & 63.2 \\ \hline \hline
Global(IC15) & 78.5 & 68.6 & 76.6 & 70.2 & 78.6 & 71.4 & 78.5 & 70.4 & 78.4 & 69.6 \\
Local(IC15) & 78.5 & 68.9 & 78.7 & 71.3 & 78.8 & 72.6 & 78.6 & 71.5 & 78.5 & 69.8 \\ \hline
\end{tabular}%
}
\vspace{-0.3cm}
\end{table}
\vspace{-0.1cm}
\begin{table}[!hbt]
\centering
\caption{Relative contribution (WRA) of three design choices behind training Feedback Network for iterative prediction: (a) Regularisation constraint via $L_{C}$ (Eqn. \ref{feedback_loss}), (b) Modified variational lower bound for feedback network accommodating $F_{aux}$, (c) Using $\mathcal{G}$  to capture the prior-knowledge from error distribution.}
\label{tab:my-table7}
\resizebox{1.\columnwidth}{!}{%
\begin{tabular}{c|c|c|c|c|c|c}
\hline
Constraint & Auxiliary & Use Error & \multicolumn{2}{c|}{CS Setup} &  \multicolumn{2}{c}{DS Setup} \\ \cline{4-7}
$L_{C}$ & Decoder $F_{aux}$ & Dist. $\mathcal{G}$ & IC15 & IAM & IC15 & IAM \\ \hline 
 \checkmark& - & - & 76.8 & 85.3 & 69.6 & 59.8 \\ 
 - & \checkmark &  \checkmark& 76.6 & 84.5 & 68.7 & 59.3 \\ 
 \checkmark & \checkmark & - & 76.9 & 85.3 & 70.1 & 59.9 \\ 
  \checkmark & \checkmark & \checkmark & 78.8 & 87.5 & 72.6 & 64.8 \\ 
 \hline \end{tabular}%
  \vspace{-0.25cm}
}
\end{table}

\noindent \textbf{Feedback module vs Language Model:} We could have substituted our iterative approach towards refining text prediction with a Language model(LM). For fairness, we use a state-of-the-art RNN-LM \cite{GuoWeiss2019} trained from text corpus (librispeech) at character level \cite{WinNT} that aims to predict the next likely character. This could be fused with the text recognition decoder using two state-of-the-art methods introduced in \cite{gulcehre2015using} via \textit{Shallow Fusion} (weighted sum of predicted scores) and \textit{Deep Fusion} (fusing their hidden states). Our method performs better in both CS and DS setups in comparison to these LM integrations (Table \ref{table6}). The limited performance of off-the-shelf LM  \cite{kang2019candidate} can be attributed to: (i) LM is mostly used in speech recognition tasks, where data is present at sentence-level which provides enough context. For distract word recognition (our focus) however, LM cannot harness such extent of context information. (ii) The LM corpus is significantly different from that used for training word-image recognition system. This leads to a biased incorrectness \cite{GuoWeiss2019}.
(iii) LM being an independent post processing step, not only ignores rich visual features from the input image, but is also unaware of the error distribution of the model. On the contrary, our model revisits the rich visual features iteratively after every prediction, considering the error distribution while training. Furthermore to align with the evaluation standards for \emph{unconstrained word recognition} we cite all results in our work using greedy decoding only \textbf{--} no LM based post-processing.
\begin{table}[!hbt]
 \vspace{-0.20cm}
\footnotesize
\centering
\caption{Complexity and speed analysis against no. of parameters and flops (Multiply-Add), for both individual component (left) and varying no, of iterations (right), using CPU time  during inference.}
 \label{table5}
    \resizebox{1.0\columnwidth}{!}{%
    \begin{tabular}{ccccccc}
        \cline{1-4} \cline{6-7}
         Network & Parameters & Multiply-Add & CPU && Iteration & CPU \\
         \cline{1-4} \cline{6-7}
         $T_{A}$ & 5.4M & 1461M & 8.27ms && t=0 & 28.64ms \\
         $T_{B}$ & 23.7M & 1428M & 12.82ms && t=1 & 43.69ms \\
         F & 24.4M & 1897M & 2.19ms && t=2 & 58.66ms \\
         \cline{1-4} \cline{6-7}
    \end{tabular}
    }
\end{table}
\vspace{-0.4cm}

\begin{table}[H]
\centering
 \vspace{-0.10cm}
\caption{Comparison with different LM integration methods.}
\label{table6}
\resizebox{1.\columnwidth}{!}{%
\begin{tabular}{c|cc|cc|cc|cc|cc}
\hline
\multirow{3}{*}{Methods} & \multicolumn{10}{c}{Conventional Setup \& Disjoint Setup} \\ \cline{2-11} 
  &  \multicolumn{2}{c|}{IIIT5K} & \multicolumn{2}{c|}{IC15} & \multicolumn{2}{c|}{CUTE80} & \multicolumn{2}{c|}{IAM} & \multicolumn{2}{c}{RIMES} \\ \cline{2-11}
 & CS & DS & CS & DS & CS & DS & CS & DS & CS & DS \\ \hline
Shi \etal \cite{ShiBaiPAMI2019} & 93.2 & 84.3 & 75.9 & 65.7 & 79.3 & 61.6 & 82.36 & 54.4 & 88.9 & 59.7 \\
\cite{ShiBaiPAMI2019} + Shallow & 93.3 & 84.3 & 75.9 & 65.7 & 79.3 & 61.6 & 82.30 & 54.3 & 88.7 & 59.7 \\
\cite{ShiBaiPAMI2019}  + Deep & 93.5 & 85.6 & 76.5 & 67.4 & 81.2 & 62.9 & 83.67 & 57.5 & 89.9 & 63.6 \\
\cite{ShiBaiPAMI2019} \textbf{+ CVAE-Feed.} (t=2) & 94.9 & 90.8 & 78.8 & 72.6 & 82.9 & 66.1 & 87.5 & 64.8 & 92.7 & 70.5 \\
\hline 
\end{tabular}%
} 
\vspace{-0.2cm}
\end{table}

\vspace{-0.3cm}
\section{Conclusion}
\vspace{-0.1cm}
Here we have proposed a novel \emph{iterative approach} towards text recognition. Using a conditional variational autoencoder (CVAE) as a feedback module, the knowledge of predicted character sequences is passed from the previous iteration, into the main recognition network, improving subsequent predictions.
Our feedback network learns to use the \emph{error distribution} among multiple character sequences that are closely related to a candidate word. Experiments on various STR and HTR datasets show our network to outperform others on the conventional setting, and more significantly on the more practical disjoint (unseen) setting. 


{\small
\bibliographystyle{ieee_fullname}
\bibliography{egbib}

\begin{thebibliography}{10}\itemsep=-1pt

\bibitem{WinNT}
End-to-end automatic speech recognition systems - pytorch implementation.
\newblock \url{https://github.com/Alexander-H-Liu/End-to-end-ASR-Pytorch}.
\newblock Accessed on: 04-03-2021.

\bibitem{baek2019wrong}
Jeonghun Baek, Geewook Kim, Junyeop Lee, Sungrae Park, Dongyoon Han, Sangdoo
  Yun, Seong~Joon Oh, and Hwalsuk Lee.
\newblock What is wrong with scene text recognition model comparisons? dataset
  and model analysis.
\newblock In {\em ICCV}, 2019.

\bibitem{BahdanauBengio2014}
Dzmitry Bahdanau, Kyunghyun Cho, and Yoshua Bengio.
\newblock Neural machine translation by jointly learning to align and
  translate.
\newblock In {\em ICLR}, 2015.

\bibitem{bhunia2019handwriting}
Ayan~Kumar Bhunia, Abhirup Das, Ankan~Kumar Bhunia, Perla Sai~Raj Kishore, and
  Partha~Pratim Roy.
\newblock Handwriting recognition in low-resource scripts using adversarial
  learning.
\newblock In {\em CVPR}, 2019.

\bibitem{biten2019scene}
Ali~Furkan Biten, Ruben Tito, Andres Mafla, Lluis Gomez, Mar{\c{c}}al
  Rusi{\~n}ol, Ernest Valveny, CV Jawahar, and Dimosthenis Karatzas.
\newblock Scene text visual question answering.
\newblock In {\em CVPR}, 2019.

\bibitem{bowman2015generating}
Samuel~R. Bowman, Luke Vilnis, Oriol Vinyals, Andrew Dai, Rafal Jozefowicz, and
  Samy Bengio.
\newblock Generating sentences from a continuous space.
\newblock In {\em CoNLL}, 2016.

\bibitem{CarreiraMalikCVPR2016}
Jo{\~{a}}o Carreira, Pulkit Agrawal, Katerina Fragkiadaki, and Jitendra Malik.
\newblock Human pose estimation with iterative error feedback.
\newblock In {\em CVPR}, 2016.

\bibitem{chen2018iterative}
Xinlei Chen, Li-Jia Li, Li Fei-Fei, and Abhinav Gupta.
\newblock Iterative visual reasoning beyond convolutions.
\newblock In {\em CVPR}, 2018.

\bibitem{ChengICCV2017}
Zhanzhan Cheng, Fan Bai, Yunlu Xu, Gang Zheng, Shiliang Pu, and Shuigeng Zhou.
\newblock Focusing attention: Towards accurate text recognition in natural
  images.
\newblock In {\em ICCV}, 2017.

\bibitem{AONRecogCVPR2018}
Zhanzhan Cheng, Yangliu Xu, Fan Bai, Yi Niu, Shiliang Pu, and Shuigeng Zhou.
\newblock Aon: Towards arbitrarily-oriented text recognition.
\newblock In {\em CVPR}, 2018.

\bibitem{GravesICML2006}
Alex Graves, Santiago Fern\'{a}ndez, Faustino Gomez, and J\"{u}rgen
  Schmidhuber.
\newblock Connectionist temporal classification: Labelling unsegmented sequence
  data with recurrent neural networks.
\newblock In {\em ICML}, 2006.

\bibitem{gulcehre2015using}
Caglar Gulcehre, Orhan Firat, Kelvin Xu, Kyunghyun Cho, Loic Barrault, Huei-Chi
  Lin, Fethi Bougares, Holger Schwenk, and Yoshua Bengio.
\newblock On using monolingual corpora in neural machine translation.
\newblock {\em arXiv preprint arXiv:1503.03535}, 2015.

\bibitem{GuoWeiss2019}
Jinxi Guo, Tara~N Sainath, and Ron~J Weiss.
\newblock A spelling correction model for end-to-end speech recognition.
\newblock In {\em ICASSP}, 2019.

\bibitem{SynthText}
Ankush Gupta, Andrea Vedaldi, and Andrew Zisserman.
\newblock Synthetic data for text localisation in natural images.
\newblock In {\em CVPR}, 2016.

\bibitem{HanCorrection}
Na-Rae Han, Martin Chodorow, and Claudia Leakcock.
\newblock Detecting errors in english article usage by non-native speakers.
\newblock {\em Natural Language Engineering}, 2006.

\bibitem{huang2017arbitrary}
Xun Huang and Serge Belongie.
\newblock Arbitrary style transfer in real-time with adaptive instance
  normalization.
\newblock In {\em ICCV}, 2017.

\bibitem{HuhZhangCVPR2019}
Minyoung Huh, Shao-Hua Sun, and Ning Zhang.
\newblock Feedback adversarial learning: Spatial feedback for improving
  generative adversarial networks.
\newblock In {\em CVPR}, 2019.

\bibitem{Jaderberg2015}
Max Jaderberg, Andrea Simonyan, Karen~Vidaldi, and Andrew Zisserman.
\newblock Deep structured output learning for unconstrained text recognition.
\newblock In {\em ICLR}, 2015.

\bibitem{jaderberg2014synthetic}
Max Jaderberg, Karen Simonyan, Andrea Vedaldi, and Andrew Zisserman.
\newblock Synthetic data and artificial neural networks for natural scene text
  recognition.
\newblock In {\em NeurIPS Deep Learning Workshop}, 2014.

\bibitem{JaderbergARXIV2014}
Max Jaderberg, Karen Simonyan, Andrea Vedaldi, and Andrew Zisserman.
\newblock Reading text in the wild with convolutional neuralnetworks.
\newblock {\em IJCV}, 2016.

\bibitem{jaderberg2015spatial}
Max Jaderberg, Karen Simonyan, Andrew Zisserman, and koray kavukcuoglu.
\newblock Spatial transformer networks.
\newblock In {\em NeurIPS}, 2015.

\bibitem{kang2019candidate}
Lei Kang, Pau Riba, Mauricio Villegas, Alicia Forn{\'e}s, and Mar{\c{c}}al
  Rusi{\~n}ol.
\newblock Candidate fusion: Integrating language modelling into a
  sequence-to-sequence handwritten word recognition architecture.
\newblock {\em arXiv preprint arXiv:1912.10308}, 2019.

\bibitem{ICDAR2015}
Dimosthenis Karatzas, Lluis Gomez-Bigorda, Anguelos Nicolaou, Suman Ghosh,
  Andrew Bagdanov, Masakazu Iwamura, Jiri Matas, Lukas Neumann,
  Vijay~Ramaseshan Chandrasekhar, Shijian Lu, et~al.
\newblock Icdar 2015 competition on robust reading.
\newblock In {\em ICDAR}, 2015.

\bibitem{ICDAR2013}
Dimosthenis Karatzas, Faisal Shafait, Seiichi Uchida, Masakazu Iwamura,
  Lluis~Gomez i Bigorda, Sergi~Robles Mestre, Joan Mas, David~Fernandez Mota,
  Jon~Almazan Almazan, and Lluis~Pere De~Las~Heras.
\newblock Icdar 2013 robust reading competition.
\newblock In {\em ICDAR}, 2013.

\bibitem{kingma2013auto}
Diederik~P. Kingma and Max Welling.
\newblock Auto-encoding variational bayes.
\newblock In {\em ICLR}, 2014.

\bibitem{LeeCVPR2016}
Chen-Yu Lee and Simon Osindero.
\newblock Recursive recurrent nets with attention modeling for {OCR} in the
  wild.
\newblock In {\em CVPR}, 2016.

\bibitem{li2019show}
Hui Li, Peng Wang, Chunhua Shen, and Guyu Zhang.
\newblock Show, attend and read: A simple and strong baseline for irregular
  text recognition.
\newblock In {\em AAAI}, 2019.

\bibitem{LiMalikCVPR2016}
Ke Li, Bharath Hariharan, and Jitendra Malik.
\newblock Iterative instance segmentation.
\newblock In {\em CVPR}, 2016.

\bibitem{litman2020scatter}
Ron Litman, Oron Anschel, Shahar Tsiper, Roee Litman, Shai Mazor, and R
  Manmatha.
\newblock Scatter: selective context attentional scene text recognizer.
\newblock In {\em CVPR}, 2020.

\bibitem{moran}
Canjie Luo, Lianwen Jin, and Zenghui Sun.
\newblock Moran: A multi-object rectified attention network for scene text
  recognition.
\newblock {\em Pattern Recognition}, 90, 2019.

\bibitem{luo2020learn}
Canjie Luo, Yuanzhi Zhu, Lianwen Jin, and Yongpan Wang.
\newblock Learn to augment: Joint data augmentation and network optimization
  for text recognition.
\newblock In {\em CVPR}, 2020.

\bibitem{2Dattentional}
Pengyuan Lyu, Zhicheng Yang, Xinhang Leng, Xiaojun Wu, Ruiyu Li, and Xiaoyong
  Shen.
\newblock 2d attentional irregular scene text recognizer.
\newblock {\em arXiv preprint arXiv:1906.05708}, 2019.

\bibitem{IamDataset}
U-V Marti and Horst Bunke.
\newblock The iam-database: an english sentence database for offline
  handwriting recognition.
\newblock {\em IJDAR}, 2002.

\bibitem{IIIT5K-Words}
Anand Mishra, Karteek Alahari, and C.~V. Jawahar.
\newblock Scene text recognition using higher order language priors.
\newblock In {\em BMVC}, 2012.

\bibitem{HarisUkitaCVPR2018}
Norimichi~Ukita Muhammad~Haris, Greg~Shakhnarovich.
\newblock Deep back-projection networks for super-resolution.
\newblock In {\em CVPR}, 2018.

\bibitem{Neumann2012}
Luk{\'a}{\v{s}} Neumann and Ji{\v{r}}{\'\i} Matas.
\newblock Real-time scene text localization and recognition.
\newblock In {\em CVPR}, 2012.

\bibitem{NewellHourGlassECCV}
Alejandro Newell, Kaiyu Yang, and Jia Deng.
\newblock Stacked hourglass networks for human pose estimation.
\newblock In {\em ECCV}, 2016.

\bibitem{ng-etal-2014-conll}
Hwee~Tou Ng, Siew~Mei Wu, Ted Briscoe, Christian Hadiwinoto, Raymond~Hendy
  Susanto, and Christopher Bryant.
\newblock The {C}o{NLL}-2014 shared task on grammatical error correction.
\newblock In {\em CoNLL}, 2014.

\bibitem{park2019semantic}
Taesung Park, Ming-Yu Liu, Ting-Chun Wang, and Jun-Yan Zhu.
\newblock Semantic image synthesis with spatially-adaptive normalization.
\newblock In {\em CVPR}, 2019.

\bibitem{paszke2017automatic}
Adam Paszke, Sam Gross, Soumith Chintala, Gregory Chanan, Edward Yang, Zachary
  DeVito, Zeming Lin, Alban Desmaison, Luca Antiga, and Adam Lerer.
\newblock Automatic differentiation in {PyTorch}.
\newblock In {\em NeurIPS Autodiff Workshop}, 2017.

\bibitem{perez2018film}
Ethan Perez, Florian Strub, Harm De~Vries, Vincent Dumoulin, and Aaron
  Courville.
\newblock Film: Visual reasoning with a general conditioning layer.
\newblock In {\em AAAI}, 2018.

\bibitem{poznanski2016cnn}
Arik Poznanski and Lior Wolf.
\newblock Cnn-n-gram for handwritingword recognition.
\newblock In {\em CVPR}, 2016.

\bibitem{qiao2020seed}
Zhi Qiao, Yu Zhou, Dongbao Yang, Yucan Zhou, and Weiping Wang.
\newblock Seed: Semantics enhanced encoder-decoder framework for scene text
  recognition.
\newblock In {\em CVPR}, 2020.

\bibitem{SVT-P}
Trung Quy~Phan, Palaiahnakote Shivakumara, Shangxuan Tian, and Chew Lim~Tan.
\newblock Recognizing text with perspective distortion in natural scenes.
\newblock In {\em ICCV}, 2013.

\bibitem{CUTE80}
Anhar Risnumawan, Palaiahankote Shivakumara, Chee~Seng Chan, and Chew~Lim Tan.
\newblock A robust arbitrary text detection system for natural scene images.
\newblock {\em Expert Systems with Applications}, 2014.

\bibitem{RozovskayaCorrection}
Alla Rozovskaya and Dan Roth.
\newblock Generating confusion sets for context-sensitive error correction.
\newblock In {\em EMNLP}, 2010.

\bibitem{ShamaManorAFL2018}
Firas Shama, Roey Mechrez, Alon Shoshan, and Lihi Zelnik-Manor.
\newblock Adversarial feedback loop.
\newblock In {\em ICCV}, 2019.

\bibitem{Shi2017}
Baoguang Shi, Xiang Bai, and Cong Yao.
\newblock An end-to-end trainable neural network for image-based sequence
  recognition and its application to scene text recognition.
\newblock {\em IEEE T-PAMI}, 2017.

\bibitem{ShiBai2016}
Baoguang Shi, Xinggang Wang, Pengyuan Lyu, Cong Yao, and Xiang Bai.
\newblock Robust scene text recognition with automatic rectification.
\newblock In {\em CVPR}, 2016.

\bibitem{ShiBaiPAMI2019}
Baoguang Shi, Mingkun Yang, Xinggang Wang, Pengyuan Lyu, Cong Yao, and Xiang
  Bai.
\newblock Aster: An attentional scene text recognizer with flexible
  rectification.
\newblock {\em IEEE T-PAMI}, 2018.

\bibitem{sohn2015learning}
Kihyuk Sohn, Honglak Lee, and Xinchen Yan.
\newblock Learning structured output representation using deep conditional
  generative models.
\newblock In {\em NeurIPS}, 2015.

\bibitem{wan2020vocabulary}
Zhaoyi Wan, Jielei Zhang, Liang Zhang, Jiebo Luo, and Cong Yao.
\newblock On vocabulary reliance in scene text recognition.
\newblock In {\em CVPR}, 2020.

\bibitem{WangICCV2011}
Kai Wang, Boris Babenko, and Serge Belongie.
\newblock End-to-end scene text recognition.
\newblock In {\em ICCV}, 2011.

\bibitem{wang2020scene}
Wenjia Wang, Enze Xie, Xuebo Liu, Wenhai Wang, Ding Liang, Chunhua Shen, and
  Xiang Bai.
\newblock Scene text image super-resolution in the wild.
\newblock In {\em ECCV}, 2020.

\bibitem{wei2016convolutional}
Shih-En Wei, Varun Ramakrishna, Takeo Kanade, and Yaser Sheikh.
\newblock Convolutional pose machines.
\newblock In {\em CVPR}, 2016.

\bibitem{XieNG2016}
Ziang Xie, Anand Avati, Naveen Arivazhagan, Dan Jurafsky, and Andrew~Y Ng.
\newblock Neural language correction with character-based attention.
\newblock {\em arXiv preprint arXiv:1603.09727}, 2016.

\bibitem{xu2020machines}
Xing Xu, Jiefu Chen, Jinhui Xiao, Lianli Gao, Fumin Shen, and Heng~Tao Shen.
\newblock What machines see is not what they get: Fooling scene text
  recognition models with adversarial text images.
\newblock In {\em CVPR}, 2020.

\bibitem{yang2019symmetry}
MingKun Yang, Yushuo Guan, Minghui Liao, Xin He, Kaigui Bian, Song Bai, Cong
  Yao, and Xiang Bai.
\newblock Symmetry-constrained rectification network for scene text
  recognition.
\newblock In {\em ICCV}.

\bibitem{ye2017zero}
Meng Ye and Yuhong Guo.
\newblock Zero-shot classification with discriminative semantic representation
  learning.
\newblock In {\em CVPR}, 2017.

\bibitem{yue2020robustscanner}
Xiaoyu Yue, Zhanghui Kuang, Chenhao Lin, Hongbin Sun, and Wayne Zhang.
\newblock Robustscanner: Dynamically enhancing positional clues for robust text
  recognition.
\newblock In {\em ECCV}, 2020.

\bibitem{ZamirMalikCVPR2017}
Amir~Roshan Zamir, Te{-}Lin Wu, Lin Sun, William~B. Shen, Jitendra Malik, and
  Silvio Savarese.
\newblock Feedback networks.
\newblock In {\em CVPR}, 2017.

\bibitem{ESIR2019}
Fangneng Zhan and Shijian Lu.
\newblock Esir: End-to-end scene text recognition via iterative image
  rectification.
\newblock In {\em CVPR}, 2019.

\bibitem{VeriSimilarECCV2018}
Fangneng Zhan, Shijian Lu, and Chuhui Xue.
\newblock Verisimilar image synthesis for accurate detection and recognition of
  texts in scenes.
\newblock In {\em ECCV}, 2018.

\bibitem{zhang2019canet}
Chi Zhang, Guosheng Lin, Fayao Liu, Rui Yao, and Chunhua Shen.
\newblock Canet: Class-agnostic segmentation networks with iterative refinement
  and attentive few-shot learning.
\newblock In {\em CVPR}, 2019.

\bibitem{DomainAdaptationCVPR2019}
Yaping Zhang, Shuai Nie, Wenju Liu, Xing Xu, Dongxiang Zhang, and Heng~Tao
  Shen.
\newblock Sequence-to-sequence domain adaptation network for robust text image
  recognition1.
\newblock In {\em CVPR}, 2019.

\bibitem{DiscourseLevelDiversity2017}
Tiancheng Zhao, Ran Zhao, and Maxine Esk{\'{e}}nazi.
\newblock Learning discourse-level diversity for neural dialog models using
  conditional variational autoencoders.
\newblock In {\em ACL}, 2017.

\bibitem{zheng2019pluralistic}
Chuanxia Zheng, Tat-Jen Cham, and Jianfei Cai.
\newblock Pluralistic image completion.
\newblock In {\em CVPR}, 2019.

\bibitem{zheng2019distraction}
Quanlong Zheng, Xiaotian Qiao, Ying Cao, and Rynson~WH Lau.
\newblock Distraction-aware shadow detection.
\newblock In {\em CVPR}, 2019.

\end{thebibliography}
}

\end{document}